\documentclass[conference]{IEEEtran}
\IEEEoverridecommandlockouts
\usepackage{cite}
\usepackage{amsmath,amssymb,amsfonts}
\usepackage{algorithmic}
\usepackage{graphicx}
\usepackage{textcomp}
\usepackage{xcolor}
\usepackage[textsize=tiny]{todonotes}
\usepackage{booktabs}
\usepackage{multirow}
\usepackage{stfloats}
\usepackage{array}
\usepackage{tabularx}
\usepackage{xurl}
\usepackage[draft]{hyperref}
\usepackage{cleveref}
\usepackage{enumitem}
\usepackage{makecell}
\usepackage{balance}
\setlist[itemize]{itemsep=1em}

\def\BibTeX{{\rm B\kern-.05em{\sc i\kern-.025em b}\kern-.08em
    T\kern-.1667em\lower.7ex\hbox{E}\kern-.125emX}}
\begin{document}

\title{
Suicidal Comment Tree Dataset: Enhancing Risk Assessment and Prediction Through Contextual Analysis\\
}

\author{\IEEEauthorblockN{Jun Li}
\IEEEauthorblockA{\textit{Department of Computing} \\
\textit{The Hong Kong Polytechnic University}\\
Hong Kong, Hong Kong \\
hialex.li@connect.polyu.hk}
\and
\IEEEauthorblockN{Qun Zhao}
\IEEEauthorblockA{\textit{Department of Language Science and Technology} \\
\textit{The Hong Kong Polytechnic University}\\
Hong Kong, Hong Kong \\
qun.zhao@connect.polyu.hk}
}

\maketitle

\begin{abstract}

Suicide remains a critical global public health issue. While previous studies have provided valuable insights into detecting suicidal expressions in individual social media posts, limited attention has been paid to the analysis of longitudinal, sequential comment trees for predicting a user's evolving suicidal risk. Users, however, often reveal their intentions through historical posts and interactive comments over time. This study addresses this gap by investigating how the information in comment trees affects both the discrimination and prediction of users' suicidal risk levels. We constructed a high-quality annotated dataset, sourced from Reddit, which incorporates users' posting history and comments, using a refined four-label annotation framework based on the Columbia Suicide Severity Rating Scale (C-SSRS). Statistical analysis of the dataset, along with experimental results from Large Language Models (LLMs) experiments, demonstrates that incorporating comment trees data significantly enhances the discrimination and prediction of user suicidal risk levels. This research offers a novel insight to enhancing the detection accuracy of at-risk individuals, thereby providing a valuable foundation for early suicide intervention strategies.

\end{abstract}


\section{Introduction}

According to the World Health Organization (WHO), suicide claims a life every 40 seconds~\footnote{\url{https://www.who.int/data/gho/data/themes/mental-health}} globally, making it a severe public health issue. Research indicates that most risk factors can be mitigated through effective interventions, suggesting that many suicide tragedies are preventable~\cite{ahmedani2025zero}. Consequently, it is crucial to develop effective tools for the early detection and identification of individuals with potential suicidal ideation to enable timely intervention. Social media platforms, particularly mental health-focused forums like Reddit, provide a valuable data source for this purpose. Users on these platforms often express their thoughts authentically, and the data is updated in real-time, making them a vital channel for the timely and effective detection of suicide risk among individuals with mental disorders. Consequently, suicidal risk detection based on social media data has garnered significant research attention.
Utilizing Interpersonal Psychological Theory of Suicide (IPTS)~\cite{joiner2005people}, theoretical models of suicidal behavior~\cite{klonsky2015three, diaz2021theoretical}, and studies on risk and protective factors~\cite{ati2021risk}, scholars have extensively explored this area from a technological perspective. For instance, they have developed machine learning (ML) algorithms aimed at detecting suicidal ideation by analyzing various data sources, primarily social media posts containing implicit or explicit expressions of such thoughts~\cite{ji2020suicidal,li2025protective}. While ML-based approaches offer considerable advantages, the quality of the training dataset is a critical determinant of their predictive performance.

Applying ML algorithms to identify users' potential suicidal risk based on their social media posts requires well-annotated datasets. However, most existing studies on suicidal ideation detection construct datasets through keyword-based searches~\cite{klimiuk2024seasonal}, which may result in incomplete data coverage~\cite{gaur2019knowledge}. Moreover, due to the brevity of social media posts and users' varying expression styles, a single post often lacks sufficient information for accurate risk level identification. Users may reveal suicidal intentions, plans, or attempt histories through historical comment trees, which contain historical posts and corresponding interactive comments. Therefore, we argue that constructing datasets that incorporate users' historical posts and comments can provide more comprehensive information for ML models to identify and predict suicidal risk. Additionally, while LLMs possess strong general knowledge, they still require task-specific fine-tuning or In-Context Learning~\cite{brown2020language} with annotated samples to enhance performance in specific domains. To address these needs, this study constructs a high-quality annotated dataset using the four-label annotation framework~\cite{li2022suicide} based on the Columbia Suicide Severity Rating Scale (C-SSRS)~\cite{posner2008columbia}, applied to users' posts and comments from Reddit.

This study aims to address two research questions: \textbf{RQ1}: How does the inclusion of comment trees affect the discrimination of users' suicidal risk levels?
\textbf{RQ2}: How does the inclusion of comment trees affect the performance of LLMs in predicting users' suicidal risk levels? To answer the research questions, we conducted a comprehensive statistical analysis of the dataset, and applied the dataset on the Qwen3-4B, Gemini-2.5-flash, and GPT-5.
By investigating these questions, we conclusively demonstrate the critical importance of incorporating comment trees information for both the discrimination and prediction of users' suicidal risk levels. Our work provides evidence that contextual data from comment trees enhances models' performance. Furthermore, we validate the applicability of the four-label annotation framework~\cite{li2022suicide} based on C-SSRS~\cite{posner2008columbia} in ML tasks. A key contribution of this study is the refinement of this theoretical framework; based on users' variety of expressions, we have specified practical indicators for each label and clarified the distinctions between them. Furthermore, we have constructed a publicly available, high-quality dataset of 500 Reddit cases. Each case contains a user’s three chronologically ordered posts and comments, along with suicidal risk labels based on our refined criteria. Finally, our comprehensive experimental results further solidify the feasibility of using machine learning for suicidal risk prediction based on social media data, thereby providing a valuable foundation for early suicidal risk intervention.

\section{Suicidal Comment Tree Dataset}
This section presents the construction of the dataset designed for suicide risk prediction research. We describe the data collection methodology, preprocessing steps, annotation criteria and process, and statistical analysis of the curated dataset. The dataset comprises Reddit posts and their corresponding comments, providing rich contextual information for the analysis and prediction of suicidal behavior patterns.

\subsection{Data Collection and Preprocessing}
Our dataset comprises posts extracted from Reddit's ``r/SuicideWatch'' subreddit, a specialized community where users share suicide-related concerns and seek support. We utilized the dataset from~\cite{li2022suicide}, which contains 139,455 posts authored by 76,186 users between January 2020 and December 2021, collected via Reddit's official API~\cite{reddit2022api}. Data preprocessing involved anonymization of personal identifiers, and content filtering to ensure ethical compliance while maintaining analytical utility. Then, we selected 1,265 users with a total of 14,632 historical posts and crawled all corresponding comments to capture the complete comment context. We extracted sequences of three consecutive posts per user with rich comments, where the first two posts contained non-empty comments. To ensure sufficient context, segments were only selected from users with a minimum of five prior posts, with a maximum of two segments per user and a minimum gap of three posts between consecutive segments. This filtering method will enable researchers to investigate the performance of suicidal risk level prediction using longer context information in further studies. As a result, we filtered 1,959 posts with comments from 379 active users. Finally, we employed a round-robin sampling strategy to select the most recent segment from each user, creating a balanced dataset of 500 unique consecutive segments as cases for annotation and analysis. Each case includes two components: the current post and two historical comment trees. Unlike the historical comment trees, the current post refers to the user’s most recent post, collected without any community interactive comments.

\subsection{Annotation Framework}

We adopt a four-label (indicator, ideation, behavior, and attempt)~\cite{li2022suicide} that was established based on the Columbia Suicide Severity Rating Scale (C-SSRS)~\cite{posner2008columbia}, adapted for the context of social media. We have specified practical indicators for each label and clarified the distinctions between them based on users' variety of expressions. Table~\ref{tab:criteria} displays the definition and examples of the each label.


\begin{table*}[htbp]
\centering
\caption{Criteria of Different Suicidal Risk Levels}
\label{tab:criteria}
\begin{tabular}{m{0.5cm} m{1.5cm} m{5.5cm} m{7cm}}
\toprule
\textbf{Level} & \textbf{Label} & \textbf{Criteria} & \textbf{Examples} \\
\midrule
1 & Indicator & No explicit suicidal expression OR resolved suicidal issues. & You’re screaming, but no one can hear. I’ve closed the door to my bedroom, I’ve turned all the lights off and I’m eating my food, all alone. \\
\midrule
2 & Ideation & Explicit suicidal expression WITHOUT plan or tendency to commit suicide. & I’m in a crisis and I don’t trust any hotlines. Everyone just tells me to call a crisis line, but if I’m honest an say I’m gonna kill myself, they’ll call the cops on me and send me to a shitty hospital. \\
\midrule
3 & Behavior & Explicit suicidal expression WITH plan/tendency to commit suicide OR history of self-harm. (Note: Any mention of previous self-harm behaviors, even if not recent, should be classified as Behavior) & Probable suicide note. I don’t know when I’ll kill myself and it’s useless too but I’m just writing it here so that my family knows what I felt and went through at the end. My dear parents, friends, this is not your fault...\\
\midrule
4 & Attempt & Explicit expression of recent suicide attempt OR history of suicide attempt. (Note: Any mention of previous suicide attempts, even if not recent, should be classified as Attempt) & I just swallowed 45 [100mg] Lamictal pills. Is that enough? I also took some trazedone to fall asleep fast.\\
\bottomrule
\end{tabular}
\end{table*}

\textbf{Indicator}: This category primarily includes descriptions of personal experiences, feelings, and emotions that do not involve discussions about life and death. It also encompasses instances in which an individual explicitly expresses a will to live or describes protective factors~\cite{ati2021risk} and connectedness~\cite{klonsky2015three}, such as supportive relationships with family, friends, or community, which act as a buffer~\cite{cohen1985stress} against the development of suicidal ideation.

\textbf{Ideation}: This category is defined by the explicit expression of suicidal thoughts, including statements indicating a lack of motivation to live. It also encompasses instances where an individual fantasizes about or imagines scenarios of suicide. Crucially, at this stage, there is no evidence of a specific plan, preparation of means, or any mention of self-harm actions.

\textbf{Behavior}: According to the Interpersonal Theory of Suicide~\cite{joiner2005people}, individuals at risk for suicide can acquire the capability for suicide through engagement in self-harm and suicidal behaviors. This acquired capability is developed through repeated exposure to painful and provocative events, which serves to habituate the individual to fear of death and increase their tolerance for physical pain. Therefore, this label includes preparatory behaviors that precede suicidal acts, such as identifying specific timing, location, methods, and means, alongside actual self-harm behaviors.

\textbf{Attempt}: This category includes two scenarios: one is explicitly stating a past suicide attempt, and the other is when the poster mentions being in the process of a suicide attempt at the time of posting. For example, having ingested an overdose of medication (not for therapeutic purposes and potentially at a lethal dose), or hinting that they have been resuscitated after such an attempt.

\subsection{Data Annotation}

Two annotators were engaged to label each case. The labeling procedure is conducted as follows: First, the first historical comment tree was labeled independently, denoted as H1. Then, the second historical comment tree was labeled without referring to H1, denoted as H2. After that, the same second historical comment tree was labeled again, this time taking H1 into account; this label was denoted as H1+H2. Similarly, the current post was labeled without reference to any historical context, denoted as C, and then labeled again, incorporating information from two historical comment trees, denoted as H1+H2+C.

All labels were annotated and discussed between the two annotators. Cases with ambiguous expressions were marked as inconsistent. For example, the post ``I should have warned myself. I should have yelled at myself that this day would come, all too soon.'' was considered ambiguous since the purpose of ``this day'' (e.g., suicide or other negative events) was not explicit, and was therefore marked as inconsistent.

In cases where the two annotators initially disagreed, a discussion was held to reach a consensus. Such cases were recorded as disputed data. We computed inter-annotator agreement using Cohen’s Kappa~\cite{landis1977measurement}. The agreement for individual post labels (H1, H2, C) was 0.7983, indicating strong consistency. The agreement for contextualized labels that incorporate historical information (H1+H2+C) was 0.8111, further supporting the reliability of our annotation process. These results demonstrate that the labeling procedure was valid and consistent.

\begin{table*}[!htbp]
\caption{Case Example}
\label{tab:case_example}
\centering
\begin{tabular}{c p{6.5cm} c c c c}
\toprule
\textbf{Component} & \textbf{Detail} & \textbf{Item} & \textbf{Single} & \textbf{H1+H2} & \textbf{H1+H2+C} \\
\midrule
\multirow{4}{*}{\begin{tabular}{c}Comment\\Trees\end{tabular}} 
    & \textbf{Historical post 1}: Would jumping in front of a NYC subway be fast \& painless? \textbf{Comments}: (text omitted) 
& H1 & behavior & & \\
\cmidrule(lr){2-6}
& \textbf{Historical post 2}: I'm ready to go. I've taken all the pills, goodbye. \textbf{Comments}: (text omitted) 
& H2 & attempt & attempt & \\
\midrule
\multirow{2}{*}{\begin{tabular}{c}The Current\\Post\end{tabular}}
& I just swallowed 45 [100mg] lamictal pills. Is that enough? I also took some trazodone to fall asleep fast. 
& C & attempt & attempt & attempt \\
\bottomrule
\end{tabular}
\end{table*}

\section{Data Analysis}
To address RQ1 (comment trees impact on risk level discrimination), we conducted a comprehensive statistical analysis to examine the characteristics of the 500 labeled cases. Specifically, descriptive statistical analysis was employed to examine the distribution of labels and the temporal evolution of users’ suicide risk levels after incorporating comment trees information. Additionally, the Wilcoxon signed-rank test~\cite{woolson2007wilcoxon} was used to investigate the influence of comment trees information on the risk assessment of the current post. Furthermore, Spearman’s rank correlation coefficient~\cite{sedgwick2014spearman} was employed to examine the correlation between the risk level of the current post and historical comment trees information of varying scopes.

\subsection{Escalation of Suicide Risk Levels Over Time}

\begin{table}[h]
\caption{Proportion of Suicidal Risk labels across Different Length of Context}
\label{tab:label_trend_over_time}
\centering
\begin{tabular}{c c c c c}
\hline
\textbf{Item} & \textbf{Indicator} & \textbf{Ideation} & \textbf{Behavior} & \textbf{Attempt} \\
\hline
H1 & 28 & 42.8 & 22 & 7.2 \\
\hline
H1+H2 & 12 & 42 & 32.6 & 13.4 \\
\hline
H1+H2+C & 6.8 & 35.8 & 39.8 & 17.6 \\
\hline
\end{tabular}
\end{table}

\begin{figure}
    \centering
    \includegraphics[width=0.9\linewidth]{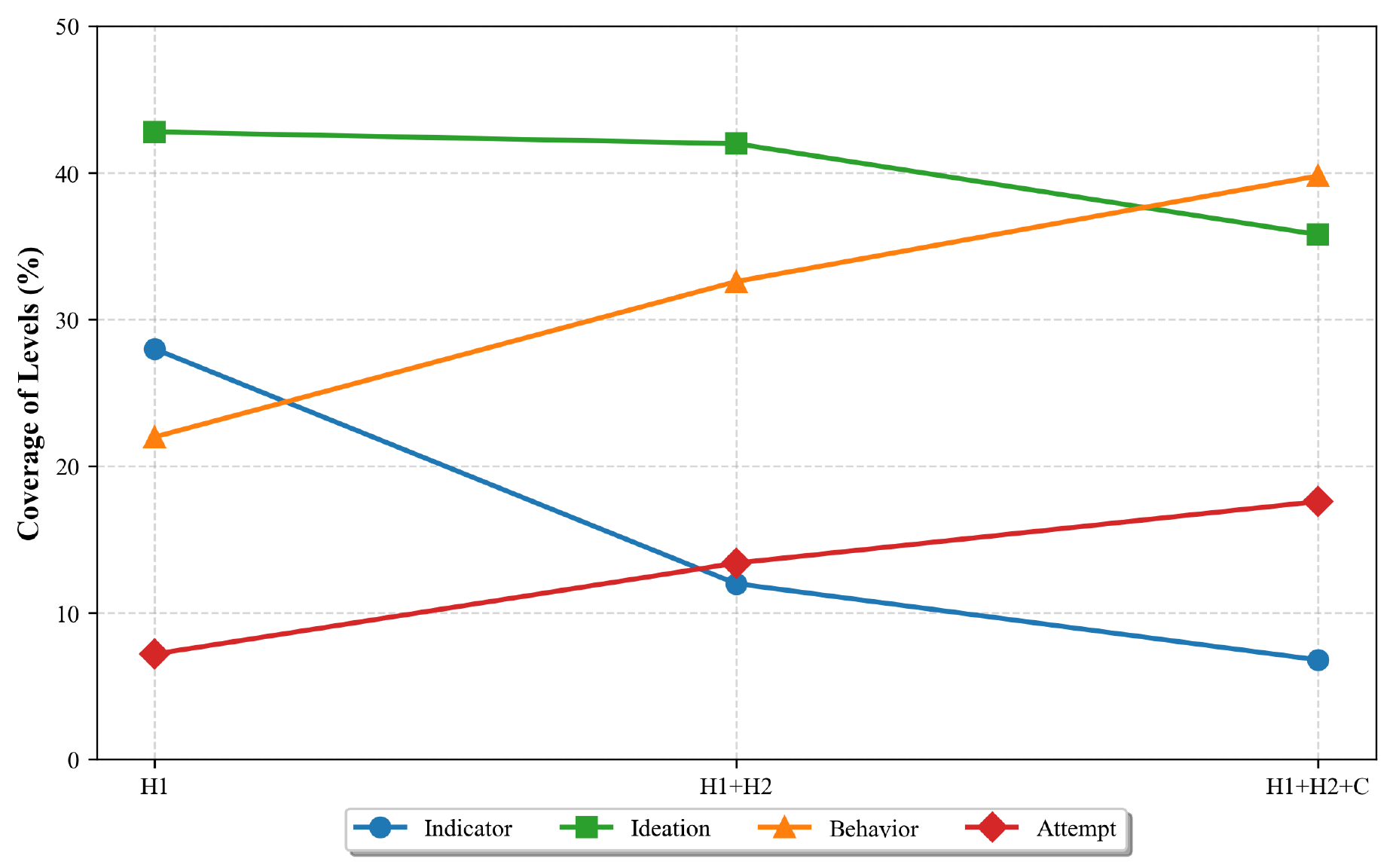}
    \caption{Suicidal Risk Levels Coverage with Incremental Historical Post Integration.}
    \label{fig:risk_levels_coverage}
\end{figure}

In Figure~\ref{fig:risk_levels_coverage}, through descriptive statistical analysis, we observe that the distribution of users' suicide risk levels demonstrates notable changes over time. From the initial stage (H1) to the intermediate phase (H1+H2), and further through the full period (H1+H2+C), the proportion of users classified under Level 1 (Indicator) and Level 2 (Ideation) gradually decreased, whereas the proportion of those in Level 3 (Behavior) and Level 4 (Attempt) consistently increased. Specifically, as shown in Table~\ref{tab:label_trend_over_time}, the percentage of users at the Indicator level declined from 28\% to 6.8\%, and those at the Ideation level decreased from 42.8\% to 35.8\%. In contrast, the proportion of users in the Behavior category rose from 22\% to 39.8\%, and those in the Attempt category increased from 7.2\% to 17.6\%. This shift in distribution suggests an overall elevation in suicidal risk among posting users over time, indicating a potential escalation in severity.

\subsection{Historical Context Significantly Impacts Suicidal Risk Assessment}

To evaluate the systematic impact of historical post information on current risk classification, we compare two scenarios: H2 versus H1+H2, and C versus H1+H2+C. The analysis revealed that incorporating historical data resulted in upward revisions of perceived risk. Specifically, 112 cases showed increased risk from H2 to H1+H2, and 171 increased from C to H1+H2+C, with no instances of risk reduction in either comparison.  Wilcoxon signed-rank tests show these shifts were statistically highly significant ($p<0.001$ in both tests). The results confirm that historical comment trees consistently and substantially impact the assessment of users'  risk level, underscoring the critical importance of longitudinal contextual information in accurate risk assessment.
These findings suggest that relying solely on a single post while ignoring the contextual information from the comment trees may lead to systematic underestimation of a user's suicide risk. Such underestimation could delay critical mental health interventions and undermine life-saving efforts. In contrast, a more comprehensive analysis that incorporates the historical comment trees provides a timely and accurate assessment of risk levels, thereby offering essential information for effective clinical intervention and decision-making.

\subsection{Strong Correlation Between Comprehensive Historical Context and User’s Current Suicidal Risk}

Correlation analysis demonstrates a strong, graded influence of the quantity and scope of historical comment trees on the discrimination of current suicidal risk levels during annotation. The suicidal risk assessment derived from the two earlier posts with comment trees (H1+H2) exhibits the strongest correlation with the full-context suicidal risk rating (H1+H2+C) ($p = 0.793$, $p < 0.001$). This is followed by the assessment based on the most recent single comment tree (H2) ($p = 0.629$, $p < 0.001$), while the earliest single comment tree (H1) shows the weakest, though still highly significant, correlation ($p = 0.535$, $p < 0.001$).
This pattern underscores the critical importance of utilizing historical context for accurate risk assessment. The stronger agreement between the combined first two posts with comments (H1+H2) and the full context (H1+H2+C) suggests that a more comprehensive view of a user's historical interactions provides a more reliable foundation for evaluating current risk. Consequently, these findings provide strong empirical support for the feasibility of employing models that integrate longer sequences of historical comment trees to enhance the prediction of a user's current suicide risk level.

\section{Experiments}
To address RQ2 (comment trees affect the performance of LLMs in predicting users' suicidal risk levels), we conducted comprehensive experiments to evaluate the impact of comment tree information on LLM performance for suicidal risk prediction. Therefore, we designed a controlled experimental setup comparing model performance across different input configurations. Specifically, we evaluated three distinct input conditions:
\begin{itemize}
\item \textbf{Posts}: utilizing only the original user posts.
\item \textbf{Posts + Comments}: incorporating both user posts and their corresponding comments.
\item \textbf{Posts + Comments + Labels}: including both posts and comments along with historical suicidal risk level labels.

\end{itemize}
This structured approach allows for precise isolation of the value added by comment trees information and historical labels to the suicidal risk prediction capability of LLMs.
\subsection{Baselines}
We evaluated the following state-of-the-art LLMs as baselines in our experimental framework:
\begin{itemize}
\item \textbf{Qwen3-4B} is an open-source LLM developed by Alibaba Cloud, offering competitive performance in contextual understanding tasks.
\item \textbf{GPT-5} is the latest generative pre-trained LLM proposed by OpenAI. It supports complex reasoning and nuanced contextual analysis.
\item \textbf{Gemini 2.5 Flash} is a powerful lightweight multimodal model released by Google, which provides fast and accurate inference with robust performance.
\end{itemize}

\subsection{Experimental Settings}
To evaluate the performance of LLMs in predicting suicide risk levels from user-generated social media content, we design a structured prompting strategy based on the Chain-of-Thought (CoT) reasoning approach~\cite{wei2022chain,wang2022self}. The prompt provides clear suicidal risk level definitions and a multi-step reasoning pathway to guide the model’s prediction process. Below shows a simplified illustration of the prompt structure used in our experiments.
\begin{quote}
\small
\textbf{Instruction}: 

\textit{You are a suicide risk assessment expert. Given a user's two historical posts with comments and their annotated risk levels in chronological order (from earlier to later), please predict the suicidal risk level of their subsequent post using the following risk level definitions:} \textit{[Definitions from Table~\ref{tab:criteria}]}

\textbf{Reasoning Steps}:

\textit{1. Read the input sentences carefully; analyze word choice, meaning, and emotional expression.}

\textit{2. Extract key information: wish to die; method, date, place, or tools for suicide; experience of suicide attempt.}

\textit{3. Based on the analysis, assess the user’s historical posts' suicidal risk level.}

\textit{4. Predict the subsequent suicidal risk level.}

\textbf{Output Format}:
\textit{Subsequent suicide risk level: indicator/ideation/behavior/attempt.}

\textbf{Example}:
\textit{[Provided example as shown in Table~\ref{tab:case_example}]}

\textbf{Input Posts with Comments and Suicidal Risk Levels (Chronological Order)}:
\textit{[Posts with comments as provided by users]}

\end{quote}

This prompt structure ensures that the model’s predictions are based on a consistent, interpretable, and clinically-informed reasoning process. The use of CoT allows the model to decompose the task into sequential steps, thereby improving the transparency and reliability of the prediction. The same prompt was applied across all evaluated LLMs under identical settings to ensure fair comparison.

\section{Results}
Table~\ref{tab:model_comparison} summarizes the weighted average precision, recall, and F1-score of three LLMs for the suicidal risk level prediction tasks with different input configurations.

\begin{table*}[htbp]
\centering
\caption{Performance Comparison of Suicide Risk Prediction Models Across Different Input Configurations, where C and L denote Comments and Label}
\label{tab:model_comparison}
\begin{tabular}{@{}l l c c c c c c@{}}
\toprule
\textbf{Model} & \textbf{Input Type} & \textbf{Prec.} & \textbf{Rec.} & \textbf{F1} & \textbf{Prec. $\mathbf{\Delta}$} & \textbf{Rec. $\mathbf{\Delta}$} & \textbf{F1 $\mathbf{\Delta}$} \\
\midrule
\multirow{3}{*}{Qwen3-4B}
& Posts & 0.6144 & 0.5640 & 0.5326 & -- & -- & -- \\
& Posts + C & 0.5667 & 0.5140 & 0.4694 & -0.0477 & -0.0500 & -0.0632 \\
& Posts + C + L & 0.5892 & 0.5840 & 0.5634 & -0.0252 & +0.0200 & +0.0308 \\
\addlinespace
\multirow{3}{*}{Gemini-2.5-Flash}
& Posts & 0.6417 & 0.6260 & 0.6300 & -- & -- & -- \\
& Posts + C & 0.6604 & 0.6520 & 0.6546 & +0.0187 & +0.0260 & +0.0246 \\
& Posts + C + L & 0.6755 & 0.6600 & 0.6596 & +0.0338 & +0.0340 & +0.0296 \\
\addlinespace
\multirow{3}{*}{GPT-5}
& Posts & 0.6283 & 0.5640 & 0.5675 & -- & -- & -- \\
& Posts + C & 0.6374 & 0.5820 & 0.5950 & +0.0091 & +0.0180 & +0.0275 \\
& Posts + C + L & 0.6509 & 0.5700 & 0.5572 & +0.0226 & +0.0060 & -0.0103 \\
\bottomrule
\end{tabular}
\end{table*}

\subsection{The Impact of Comment trees Information on Model Performance}
By comparing the experimental results of three different input configurations, we observe that the influence of comment trees information on suicide risk prediction models exhibits significant model-specific variability. The Gemini-2.5-Flash model demonstrates consistent performance improvement after incorporating comment information, with its weighted F1 score increasing from the baseline of 0.6300 to 0.6546, a relative gain of 2.46\%. This suggests that the model can effectively leverage additional semantic information and social interaction patterns from comments to enhance risk prediction capabilities. In contrast, the Qwen3-4B model exhibits a notable performance decline when only comment information is added, with the weighted average F1 score dropping from 0.5326 to 0.4694—a decrease of 6.32\%. This phenomenon may reflect the limitations of open-source models in processing complex social media texts and long-sequence information, as well as their sensitivity to noisy data. The GPT-5 model shows a relatively stable improving trend after incorporating comment information, with slight gains across all metrics, demonstrating the advantage of commercial models in information integration.

\subsection{The Compensatory Effect of Historical Label Information}
The introduction of historical risk level labels has brought significant compensatory and enhancing effects to model performance, a phenomenon that varies across models but is generally positive. For the previously underperforming Qwen3-4B model, historical labels contributed to a notable performance recovery, increasing its weighted average F1 score substantially from 0.4694 to 0.5634. This not only offset the negative impact caused by comment information but also surpassed the original baseline performance. Such a strong compensatory effect suggests that the prior knowledge provided by historical labels plays a critical role in guiding the model to understand user behavior patterns. Gemini-2.5-Flash continued to maintain its leading performance, reaching the highest weighted F1 average score of 0.6596 with historical labels—a 2.96\% improvement over the baseline. However, it is worth noting that GPT-5 exhibited a decline in F1 score after incorporating historical labels, dropping from 0.5950 to 0.5572. This may indicate potential information conflicts when processing multi-source heterogeneous data. Overall, historical labels serve as valuable contextual information, providing useful predictive cues for most models.

\section{Ethical Consideration}
The task of suicidal risk prediction based on social media data introduces significant ethical challenges that must be rigorously addressed. Privacy and confidentiality are of critical importance, as suicide-related posts contain highly sensitive personal information that must be protected through stringent data security measures. At last, it is critical to acknowledge that automated suicidal risk prediction systems carry inherent limitations and should not be considered a replacement for professional mental healthcare. These systems may generate inappropriate responses during crisis situations or fail to recognize complex clinical nuances that require human expertise.


\section{Conclusion}
This study addresses the critical need for early suicide detection by investigating how contextual comment information enhances machine learning performance. Our findings conclusively demonstrate that including comment trees significantly improves both suicidal label identification and LLM predictive performance. We refined the C-SSRS-based four-label framework by specifying practical linguistic indicators for each suicidal risk level and constructed a high-quality dataset of 500 manually annotated Reddit cases with comment threads. This dataset is publicly available to support computational mental health research.

While our study shows promise, our study has limitations including platform-specific data (Reddit only) and dataset size constraints. Future work should validate these methods across multiple social media platforms, develop models capable of processing longer post-comment contexts and bridge the gap between research and practice by integrating these tools into clinical workflows to support mental health professionals in early identification and intervention for at-risk individuals.

\balance
\bibliographystyle{IEEEtran}
\bibliography{bibs}

\end{document}